# Estimation of excess air coefficient on coal combustion processes via gauss model and artificial neural network

Sedat Golgiyaz [a],*, Muhammed Fatih Talu [b], Mahmut Daşkın [c], Cem Onat [d]

[a] *Department of Computer Engineering, Bingol University, 12000 Bingol, Turkey*
[b] *Department of Computer Engineering, Inonu University, 44280 Malatya, Turkey*
[c] *Department of Mechanical Engineering, Inonu University, 44280 Malatya, Turkey*
[d] *Department of Airframe and Power-plant, Civil Aviation High School, Firat University, 23180 Elazig, Turkey*

**Abstract** It is no doubt that the most important contributing cause of global efficiency of coal fired thermal systems is combustion efficiency. In this study, the relationship between the flame image obtained by a CCD camera and the excess air coefficient ($\lambda$) has been modelled. The model has been obtained with a three-stage approach: 1) *Data collection and synchronization*: Obtaining the flame images by means of a CCD camera mounted on a 10 cm diameter observation port, $\lambda$ data has been coordinately measured and recorded by the flue gas analyzer. 2) *Feature extraction:* Gridding the flame image, it is divided into small pieces. The uniformity of each piece to the optimal flame image has been calculated by means of modelling with single and multivariable Gaussian, calculating of color probabilities and Gauss mixture approach. 3) *Matching and testing:* A multilayer artificial neural network (ANN) has been used for the matching of feature−$\lambda$.

## 1. Introduction

For many countries, the major energy demand is fulfilled from the conventional energy resources like coal, petroleum and natural gas. In future projections, fossil fuels will continue to maintain their importance [1,2]. The world energy consumption according to the energy source and coal consumption according to the regions are given in Fig. 1. Accordingly, coal will continue to be the second largest source of energy in the world by 2030, following oil and other liquids. It is also seen from Fig. 1 that the total coal consumption in non-OECD countries has been increased. On the other hand, with the increase of coal consumption year by year, the usage of coal and other fossil fuels poses a challenge to the life and environment [3].

In recent years, engineers have a motivation for improving efficiency of coal combustion systems due to higher fuel costs, occasionally limited fuel availability, higher environmental concerns and legal limitations. Analyzing the flue gas and per-forming necessary interventions to the thermal system, efficiency of combustion is improved by closed loop control approach [4]. In order to achieve the highest combustion efficiency, complete combustion should occur. Complete combustion can only be achieved under favorable conditions. These conditions are appropriate fuel air ratio, suitable turbulence, suitable temperature and sufficient time period. The combustion process is extremely complicated, because combustion condition is very changeable [5]. When it changes, the fuel air ratio will deviate from the optimal value; correspondingly, the combustion system will deviate from the optimal state.

* Corresponding author. Tel.: +90426-2160012-1928.
E-mail address: sedatg@bingol.edu.tr (S. Golgiyaz).
Peer review under responsibility of Faculty of Engineering, Alexandria University.

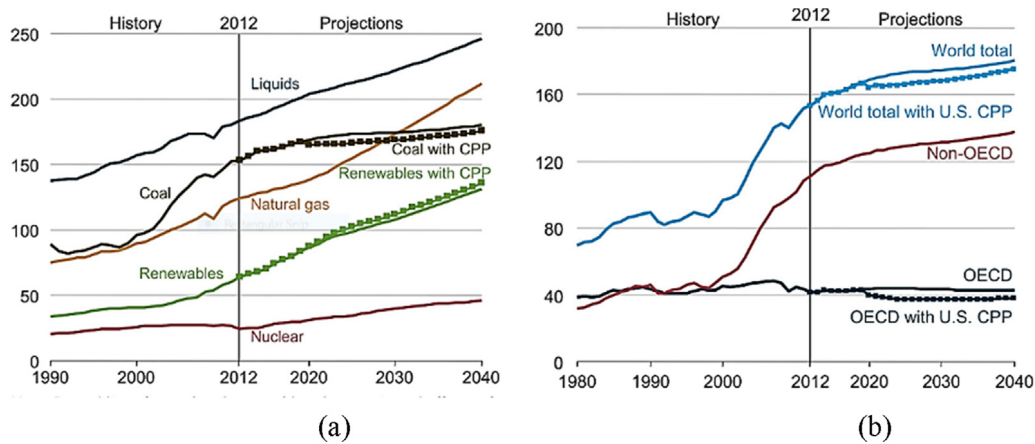

Fig. 1  World (a) energy consumption by source and (b) coal consumption by region [1].

The $\lambda$ is the most important parameter featuring the effectiveness of combustion. It also represents an ideal value for maximization of combustion effectiveness. This is very precious for combustion control. The $\lambda$ is conventionally measured by means of flue gas analyzer devices. The flue gas analyzer devices chemically and physically measure and calculate many parameters of the combustion [6]. Flue gas analyzer does not matter, it is faced a certain time delay if the combustion efficiency is determined by the data from the flue. The time delay in the system not only makes the control design process more complex, but also negatively affects the controller performance in practice [7,8]. On the contrary, the flame images can immediately reflect the current status of the combustors and so they are crucial in perspective of the combustion control because of decreasing structural dead time of the combustion process in as much as the dead time affects adversely disturbance rejection performance of closed loop control systems. The proposed CCD camera system automatically controls the combustion and allows the user to monitor the burning process at the same time. Due to the ability of the flame images, there is abundance of theory on flame image processing technology. [9] and [10] studies represent an overview of this topic. CCD cameras have been frequently used in flame imaging studies because these are widespread in the industry and these problems and superior aspects are well known [11,12,21–28,13–20]. Various laser-based visualization techniques have been also used to investigate the combustion process [29,30].

In applications carried out with flame image, image intensity and spectral analysis of the image have been widely used to express combustion characteristics. In studies which carry out with flame intensity; flame image histogram [21,22] and 4 statistical moments [12–14,25,31] of the image; Mean, standard deviation, kurtosis and skewness are commonly used features. The most sensitive parameters to temperature and air/fuel ratio are reported to be brightness (mean gray value), fluctuation amplitude (standard deviation) and flicker (weighted average oscillation frequency) [25]. However, it is emphasized that new experimental tests are needed to confirm the detected dependencies. Spectral analysis [14,25,32] and grayscale image gradient magnitude values [8] are often used together with statistical moment techniques.

In the studies conducted for the analysis of the combustion process, radiant energy naturally emitted from the flame has been widely used by researchers [13,21,27,31,33–35]. Radiant energy signal (RES) is one of the important informative parameters. It is simply obtained from a flame image, which can reflect the rapid variation in coal amount, and can also sense the transient change of coal quality. RES can be obtained by means of different approaches. For example, RES is an average gray value in the flame image in [13,31]. RES can also be a light intensity signal of the combustion chamber [27]. Researchers explained the flame temperature distribution and the RES are sensitive to change in the combustion of the boiler [33]. Also, RES and combustion rate can be associated [34]. The relationship between fuel consumption rate and heat absorption rate can be achieved with RES [35]. Besides, Talu et al. have shown that RES is not as good as competence for prediction of $\lambda$ [21]. A method based on the co-occurrence matrix for the estimation of $\lambda$ from flame images taken through a CCD camera is proposed in [21]. With their method, much more accuracy has been gained than RES based methods. However, the accuracy of the estimation system proposed in [21] is insufficient to use in a closed loop control system.

When the studies performed with flame image are evaluated in general, the proposed feature extraction methods are not sufficient to determine the combustion characterization. Despite the proposed different approaches, monitoring of the combustion process is still under development. In addition to this, significant progresses are expected by the combustion industry to facilitate the understanding of combustion processes. In this perspective, this paper represents a new approach for the estimating of the $\lambda$. As can be seen from the experimental results presented in this study, better results were obtained with the proposed method. The proposed approach consists of three steps. In the first step, the model parameters of the ideal flame images have been calculated. In second step, flame images have been converted to features by means of the obtained model parameters. And finally, the features are matched with λ.

The paper is organized as follows: The experimental burner system and data acquisition system are described in Section 2. Uniformity of the ideal flame form is presented in Section 3. ANN system is described in Section 4. Experimental results are given in Section 5. Concluding remarks are given in Section 6.

## 2. Experimental setup and data acquisition

In the experimental system, an auto-loading nut-coal burner has been used. The capacity of the small-scale burner is 85000 kcal/h. The window position on the boiler is of course important for a clear view of the combustion process inside. A circular window with a diameter of 10 cm was opened on the boiler. The position of this window is carefully determined. The combustion process was visualized with the aid of a camera positioned approximately 2 cm away from this window. Positioning and adjusting the angle are made by an expert user according to pre-defined information through the software. An exemplary image obtained is shown in Fig. 2. The circular region resembling the sun refers to the burning area. If the sun surface is spotless and has a dark golden yellow color, it is observed that the combustion efficiency is maximum. The loss of its golden yellow color and the appearance of black spots cause a decrease in combustion efficiency. Therefore, when examining the sun-like flame image is divided into equal-sized regions and the color characteristics of each region are evaluated separately.

The proposed image-based approach cannot be used to track old coal boilers without a window. However, opening a small window with a diameter of 10 cm will be enough to use the proposed approach.

The experimental system is shown in Fig. 3. Experimental data have been collected from the steady state response of the system. In other words, the transient response of the system has been neglected. While the system has been in steady state, the CCD camera images and the real time λ values obtained from the flue gas analyzer have been recorded. The real λ values are used to update the ANN weights at training phase and to measure proposed model accuracy score at training phase. The values of λ have been recorded one value per second. On the other hand, CCD camera images have been recorded two values per second. The flame display region is set to $1088 \times 1088$ pixel. During the recording period of 83 minutes, 4980 λ and 9960 flame images have been obtained. λ and image data have been simultaneously recorded by using of a computer. The cubic interpolation method has been used to calculate the missing λ values. Thus, a λ value is matched to each image. Fig. 4 shows the change of the λ value and the flame images for some λ values.

It can be seen from Fig. 4 that the different values of λ are altered in the form of flame and, the relationship between λ and the flame image form is noticeable. Mathematical modeling of this relationship is the main purpose of this study. In the following section, this relationship is modeled with the proposed original methodology.

## 3. Similarity to the ideal flame form

This section includes the modeling of the flame images at the ideal λ, the similarity of the flame images in the combustion process to the model, and how they are transformed into the feature vectors of each image by using similarity values (three different approaches).

### 3.1. Modelling

The λ values at the time of ideal burning is between 1.2 and 1.5 [36]. In the combustion process, 22 flame images with a λ value in the specified range were used to model the ideal combustion. Images were examined by separating into independent color channels (red-R, green-G, blue-B, gray-I) (see Fig. 5).

The first two moments of the brightness values in each channel are calculated as follows.

$$\mu_k = \frac{1}{MxN}\sum_{i=1}^{M}\sum_{j=1}^{N} k_{i,j} \qquad (1)$$

$$\sigma_k = \sqrt{\frac{1}{MxN}\sum_{i=1}^{M}\sum_{j=1}^{N} (k_{i,j} - \mu_k)^2} \qquad (2)$$

$k \in \{R, G, B, I\}$ color channel variable, $\sigma_k$ and $\mu_k$ are the standard deviation and mean of the brightness values in the corresponding channel. $M$ and $N$ variables express the row and column height of the flame image. The calculated torques will be used to calculate the degree to which the flame images obtained during the combustion process are similar to the ideal combustion. Also, this study investigated whether there is an important relationship between the channels. For this, [R G], [R B], [G B] and [R G B] channels were combined together. For example, the relationship between [R G] channels are defined as follows.

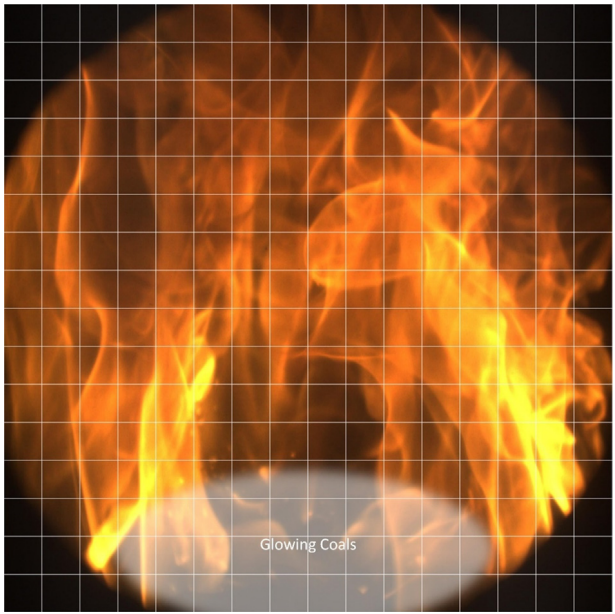

**Fig. 2** Flame image divided into regions (glowing coals are indicated by white transparent ellipse).

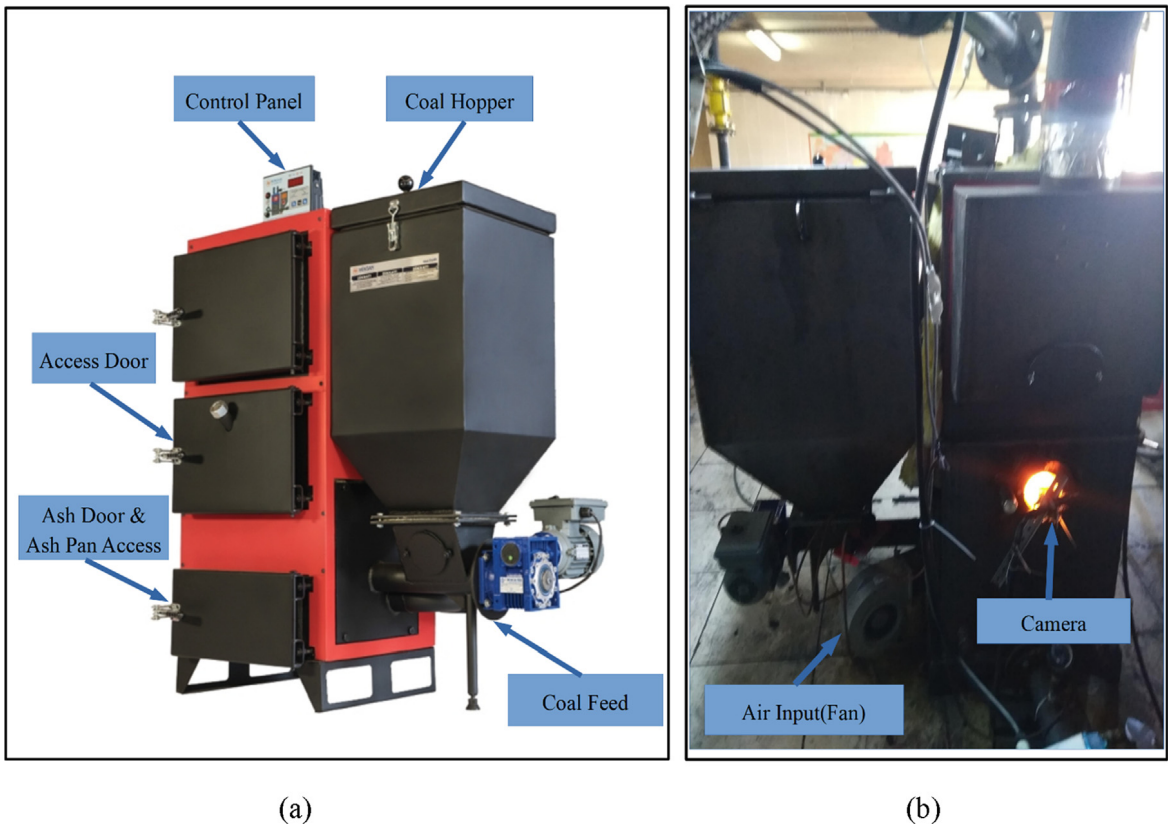

Fig. 3 The experimental burning system (a) front view (b) back view.

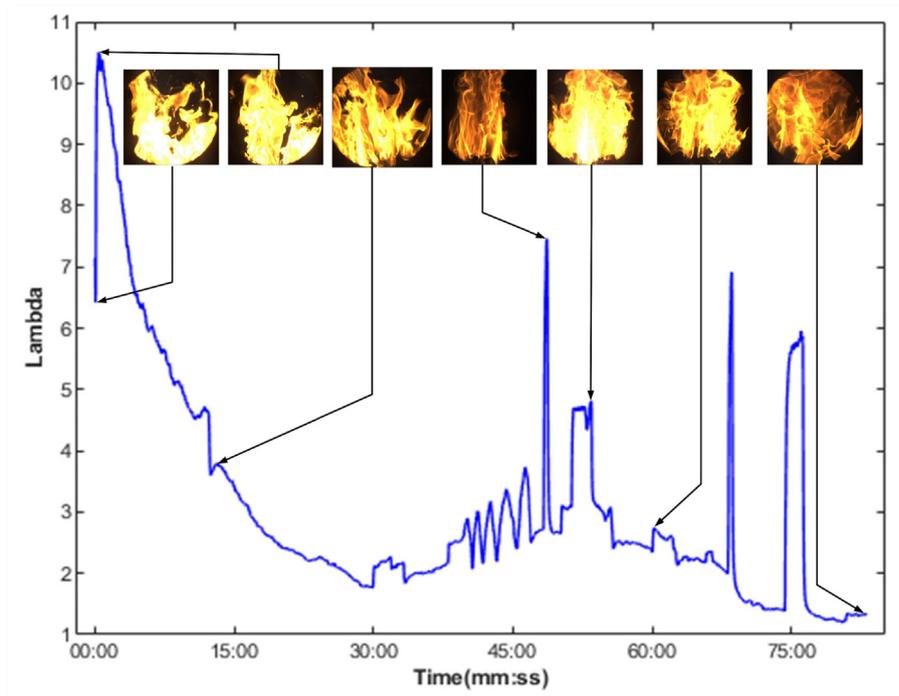

Fig. 4 Plotting λ values and some flame images.

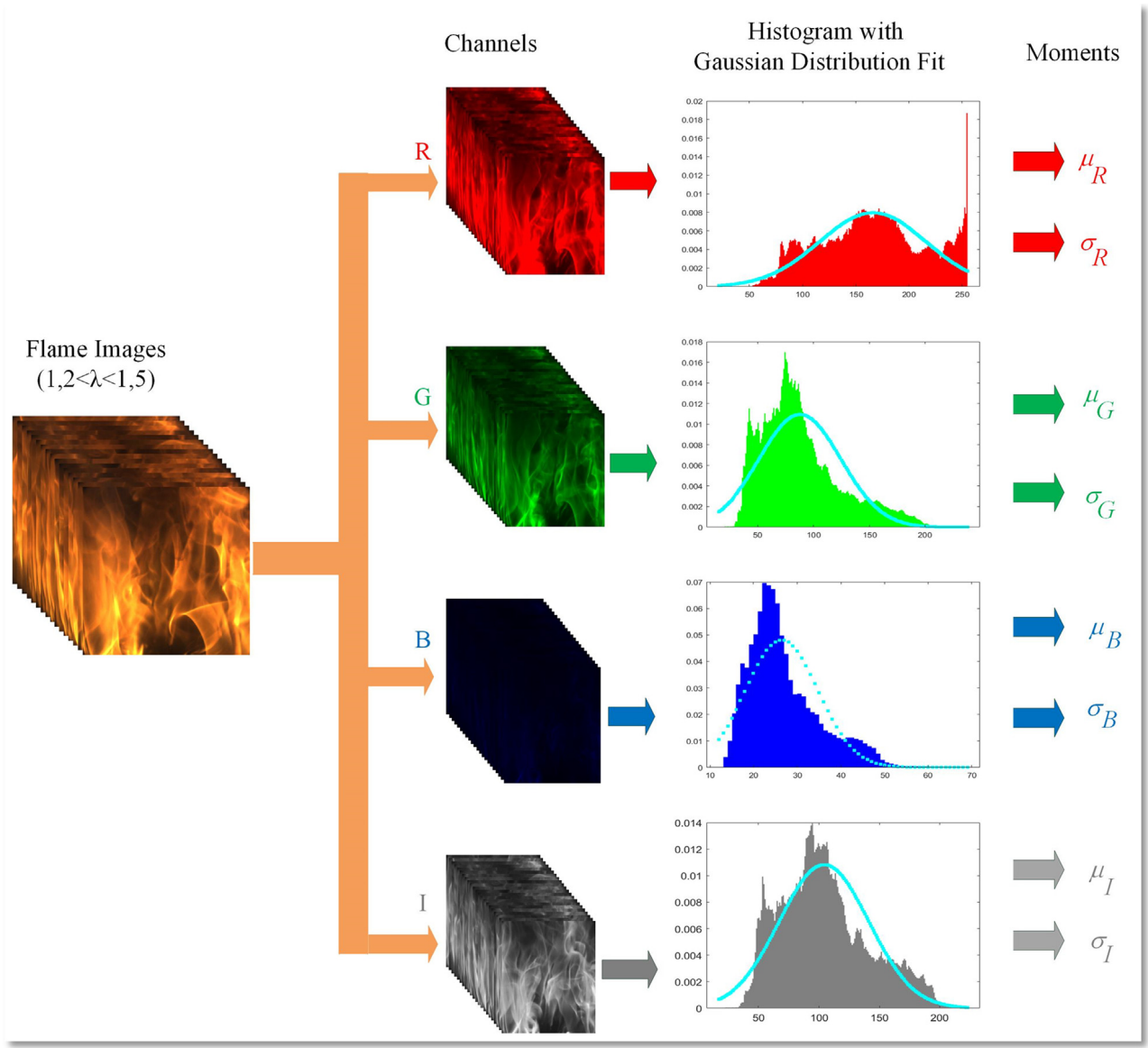

**Fig. 5** Modeling of ideal combustion.

$$\Sigma_{RG} = Cov(R, G)$$
$$= \frac{1}{MxN}\sum_{i=1}^{M}\sum_{j=1}^{N}(R_{i,j} - \mu_r)(G_{i,j} - \mu_g)^T \quad (3)$$

$$P(k_{i,j}|\mu_k, \sigma_k) = \frac{1}{\sigma_k\sqrt{2\pi}}exp\left(-\frac{(k_{i,j} - \mu_k)^2}{2\sigma_k^2}\right) \quad (4)$$

### 3.2. Pixel similarity

At this stage, the similarity of the flame images obtained during the combustion process to the ideal flame form is calculated. The similarity value can be calculated with global and local approaches. Global approaches work faster when they lose position information, but local approaches have higher transaction costs, but because of their location information, they give more accurate similarity results. In this study, the local approach has been taken and 1088x1088 sized flame images have been divided into small pieces with a size of 16x16 and the similarity of the local regions to the ideal combustion has been obtained by normal distribution function given in (4).

### 3.3. Feature vector calculation

Four different approaches were examined during the calculation of feature vectors.

**Method 1) Total Similarity**

Local window similarity is calculated as the sum of pixel similarities for each channel:

$$Feature[t, k] = \sum_{i=1}^{m}\sum_{j=1}^{n} P(k_{i,j}|\mu_k, \sigma_k) \quad (5)$$

$t$ represents the local window and $1 \leq t \leq 256$, $k$ color channel, $m$ and $n$ are the row and column height of the local window. Each region of the flame image shown in Fig. 2 is con-verted to a vector designated *Feature*[*t*, *k*]. This is done in all three-color channels of the image. The feature vector size is as many as "Window Size × Channel Count".

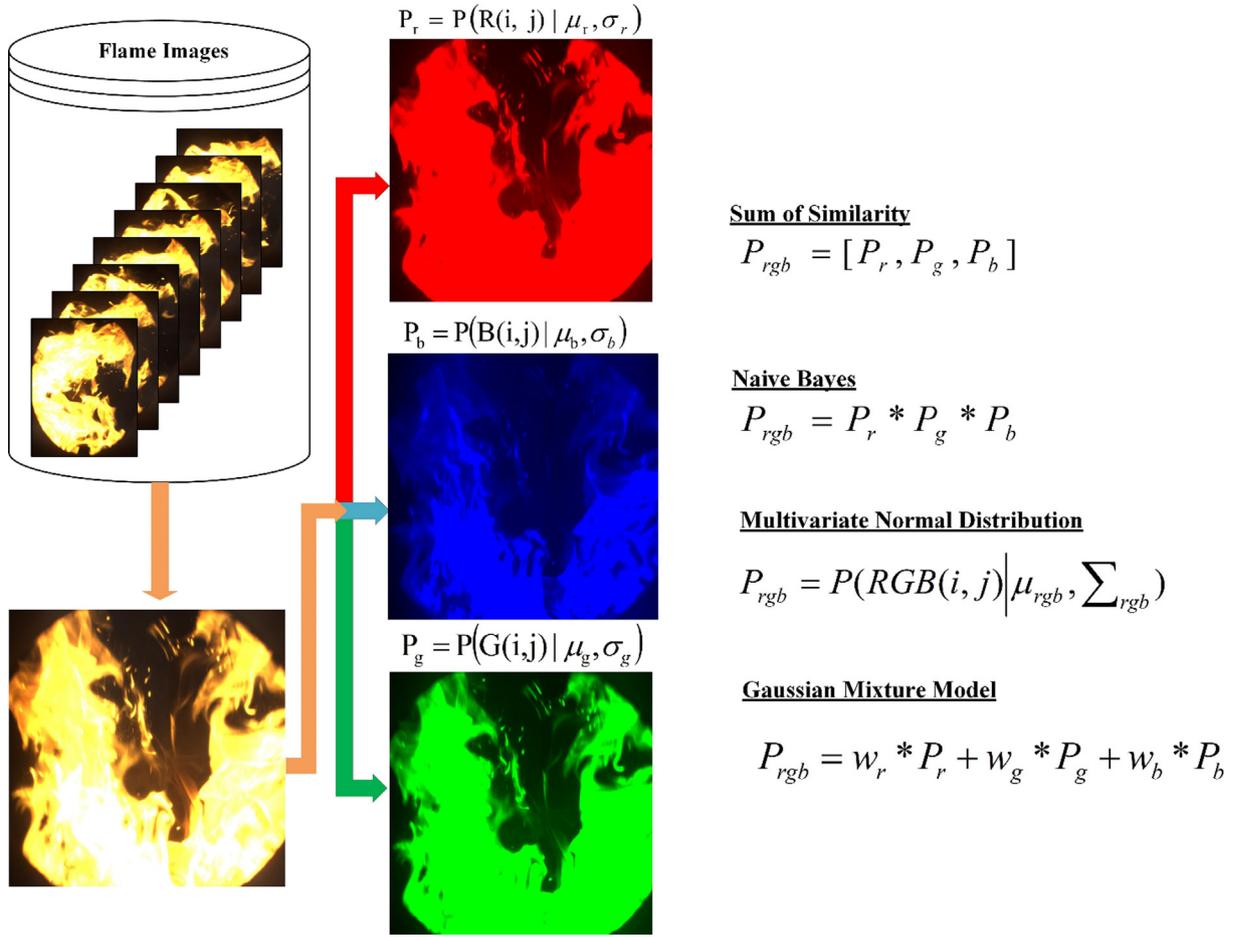

Fig. 6  Image similarity approaches.

Table 1  Separation of data in percentages for training, testing and validation.

|  | All | Training | Validation | Test |
|---|---|---|---|---|
| **Data rate** | %100 | %70 | %15 | %15 |
| **Data number** | 9956 | 6970 | 1493 | 1493 |

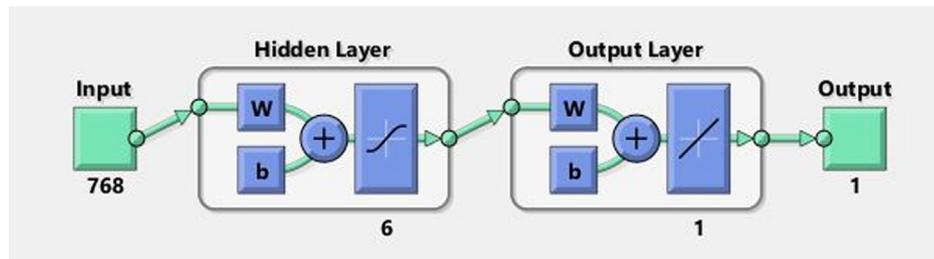

Fig. 7  Block diagram of the proposed ANN regression model.

**Method 2) Naive Bayes Similarity**

In this approach, the combination of independent color channels has been investigated. Accordingly, the similarity of the local window to the flame form is equal to the multiplication of each channel to the model. The similarity expression is given in (6).

$$Feature[t] = \prod_{k=1}^{K} P(t_k|\mu_k, \sigma_k), K \leq 3 \qquad (6)$$

**Table 2** Experimental results.

| Image Channel | Feature Vector Length | ANN Train Method | Sum of Similarity | | Naive Bayes | | Multivariate Normal Distribution | | Gaussian Mixture Model | | |
|---|---|---|---|---|---|---|---|---|---|---|---|
| | | | MSE | R | MSE | R | MSE | R | Weights $[w_R, w_G, w_B]$ | MSE | R |
| R | 256 | SCG | 0.55 | 0.9143 | - | - | - | - | - | - | - |
| G | 256 | SCG | 0.35 | 0.9463 | - | - | - | - | - | - | - |
| B | 256 | SCG | 0.54 | 0.9153 | - | - | - | - | - | - | - |
| I | 256 | SCG | 0.41 | 0.9361 | - | - | - | - | - | - | - |
| G-B | 512 | SCG | 0.34 | 0.9471 | 0.81 | 0.8702 | 0.78 | 0.8756 | 0.95-0.05 | 0.37 | 0.9424 |
| R-B | 512 | SCG | 0.32 | 0.9505 | 1.03 | 0.8305 | 0.58 | 0.9094 | 0.55-0.45 | 0.45 | 0.9301 |
| R-G | 512 | SCG | 0.33 | 0.9501 | 0.77 | 0.8772 | 0.32 | 0.9509 | 0.25-0.75 | 0.35 | 0.9468 |
| R-G-B | 768 | SCG | **0.25** | **0.9626** | 1.25 | 0.7905 | 0.93 | 0.8480 | 0.025-0.95-0.025 | 0.35 | 0.9460 |
| R-G-B-I | 1024 | SCG | 0.28 | 0.9568 | - | - | - | - | - | - | - |
| R | 256 | LM | 0.41 | 0.9383 | - | - | - | - | - | - | - |
| G | 256 | LM | 0.32 | 0.9519 | - | - | - | - | - | - | - |
| B | 256 | LM | 0.46 | 0.9295 | - | - | - | - | - | - | - |
| I | 256 | LM | 0.37 | 0.9436 | - | - | - | - | - | - | - |
| G-B | 512 | LM | 0.30 | 0.9543 | 0.66 | 0.8975 | 0.63 | 0.9015 | 0.95-0.05 | 0.33 | 0.9498 |
| R-B | 512 | LM | 0.35 | 0.9483 | 0.65 | 0.8985 | 0.54 | 0.9178 | 0.95-0.05 | 0.41 | 0.9394 |
| R-G | 512 | LM | 0.29 | 0.9578 | 0.50 | 0.9244 | 0.31 | 0.9535 | 0.05-0.95 | 0.32 | 0.9517 |
| R-G-B | 768 | LM | 0.33 | 0.9516 | 0.74 | 0.8837 | 0.59 | 0.9095 | 0.075-0.85-0.075 | 0.33 | 0.9507 |
| R-G-B-I | 1024 | LM | 0.32 | 0.9537 | - | - | - | - | - | - | - |

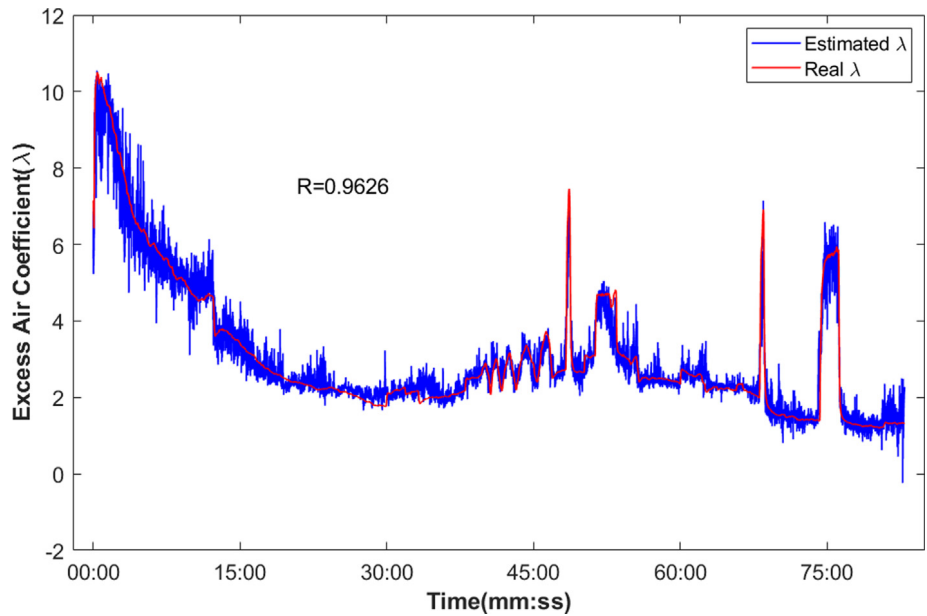

**Fig. 8** Lambda measured by the conventional gas analyzer and the proposed method.

Table 3  Image-λ matching results.

| | Method | FVL | ANN Train Method | All Perf. MSE | All Perf. R | Train Perf. MSE | Train Perf. R | Test Perf. MSE | Test Perf. R |
|---|---|---|---|---|---|---|---|---|---|
| Proposed | Sum of Similarity | 768 | SCG | 0.2453 | 0.9626 | 0.1896 | 0.9710 | 0.3751 | 0.9425 |
| Proposed | Sum of Similarity | 512 | LM | 0.2874 | 0.9578 | 0.1637 | 0.9774 | 0.5679 | 0.9128 |
| Others | Hue-Hist (0-85.230) [22] | 86 | SCG | 1.1417 | 0.8109 | 1.1387 | 0.8112 | 1.1247 | 0.8090 |
| Others | Co-occurrence [21] | 64 | SCG | 0.8793 | 0.8587 | 0.8826 | 0.8577 | 0.8540 | 0.8616 |
| Others | Blue-Hist(255) [21] | 255 | SCG | 1.3375 | 0.7743 | 1.3304 | 0.7758 | 1.3635 | 0.7678 |
| Others | PCA [20] | 2 | SCG | 1.7196 | 0.6968 | 1.7252 | 0.6983 | 1.7514 | 0.6850 |
| Others | RES-Fro-Infinity[13] | 4 | SCG | 2.1871 | 0.5843 | 2.1981 | 0.5833 | 2.1436 | 0.5886 |
| Others | Mean (RES) [12], [31] | 1 | SCG | 2.7974 | 0.4037 | 2.7924 | 0.4032 | 2.8539 | 0.4023 |
| Others | Mean, standard deviation, kurtosis and skewness [25] | 4 | SCG | 1.9012 | 0.6568 | 1.8929 | 0.6570 | 1.9394 | 0.6569 |
| Others | Mean standard deviation and flicker [14] | 3 | SCG | 1.5085 | 0.7407 | 1.5264 | 0.7381 | 1.4584 | 0.7433 |
| Others | Mean, kurtosis, skewness, standard deviation and flicker [25] | 5 | SCG | 1.5542 | 0.7315 | 1.5520 | 0.7323 | 1.5322 | 0.7293 |
| Others | Skewness standard deviation, Mean, kurtosis, magnitude[8] | 5 | SCG | 1.3966 | 0.7629 | 1.3993 | 0.7620 | 1.3835 | 0.7699 |
| Others | Hue-Hist(0-85,230)[22] | 86 | LM | 0.3951 | 0.9391 | 0.3539 | 0.9456 | 0.5099 | 0.9216 |
| Others | Co-occurrence [21] | 64 | LM | 0.5624 | 0.9174 | 0.4059 | 0.9372 | 1.3880 | 0.8721 |
| Others | Blue-Hist(255) [21] | 255 | LM | 0.7216 | 0.8855 | 0.6786 | 0.8931 | 0.8437 | 0.8664 |
| Others | PCA [20] | 2 | LM | 1.6464 | 0.7128 | 1.6363 | 0.7145 | 1.6840 | 0.7114 |
| Others | RES-Fro-Infinity[13] | 4 | LM | 1.4558 | 0.7510 | 1.4408 | 0.7534 | 1.5476 | 0.7428 |
| Others | Mean (RES) [12], [31] | 1 | LM | 2.7666 | 0.4149 | 2.7914 | 0.4152 | 2.7159 | 0.4052 |
| Others | Mean, standard deviation, kurtosis and skewness [25] | 4 | LM | 1.1056 | 0.8181 | 1.1044 | 0.8193 | 1.1293 | 0.8147 |
| Others | Mean standard deviation and flicker [14] | 3 | LM | 1.0663 | 0.8252 | 1.0729 | 0.8250 | 1.0452 | 0.8281 |
| Others | Mean, kurtosis, skewness, | 5 | LM | 0.9822 | 0.8404 | 0.9654 | 0.8428 | 1.0097 | 0.8368 |

t refers to the local window, the k color channel. The gray color channel is not used here because it is derived from other channels. In other words, it depends on other channels, it is not an independent channel. The feature vector size is up to number of windows.

**Method 3) Multivariate Normal Distribution Similarity**

Each pixel in the color flame images has a multivariate structure. Therefore, the effect of a multivariate model for the similarity calculation was investigated. Local window similarity is calculated by a multivariate normal distribution:

$$Feature[t] = \frac{1}{\sqrt{|\sum_k|(2\pi)^d}} \exp\left(-\frac{1}{2}(k_{i,j} - \mu_k)\sum\nolimits^{-1}(k_{i,j} - \mu_k)^T\right) \quad (7)$$

t refers to the local window, the k color channel, the d channel number, $\sum_k \in \{\sum_{RGB}, \sum_{RG}, \sum_{RB}, \sum_{GB}\}$ covariance variable.

**Method 4) Gaussian Mixture Model (GMM) Similarity**

Three and each of two channels of color image have been used together with GMM. Its expression is given in (8).

$$Feature[t] = \sum_{i=1}^{m}\sum_{j=1}^{n} w_k P(k_{i,j}|\mu_k, \sigma_k) \quad (8)$$

The $w_k$ channel weight coefficient, $\sum w_k = 1$. The $w_k$ values are determined as follows.

$$w_1 = 0.05 + 0.1 * b, \begin{cases} w_2 = 1 - w_1, if k = 2 \\ w_{2,3} = \frac{1-w_1}{2}, if k = 3 \end{cases}, 0 \leq b \leq 9 \quad (9)$$

The process of evaluating similarity values with four different approaches is summarized in Fig. 6.

## 4. Artificial neural network

Essentially, ANNs are mathematical models inspired by the behavior of biological nerve cells [37]. In this study, a neural network consisting of three layers, the entrance, the secret and the output layer, is used. In the training phase, weights that map input data to the output are obtained. The outputs of the input data that have been not previously used are obtained in the test phase [38]. The cost function for updating the weights is given in (10).

$$J(\theta) = \frac{1}{2C}\sum_{i=1}^{C}\left(h_\theta\left(x^{(i)}\right) - y^{(i)}\right)^2 \quad (10)$$

where $C$ is data count, $\theta$ is the weights of ANN model, $m$ is the data count, $h_\theta(.)$ is a hypothesis function for instant output of ANN model, $x^{(i)}$ is the feature vector (ANN model inputs), $y^{(i)}$ is the observed instant $\lambda$ (measured from the flue gas analyzer).

The data rate used for training, testing and verification is given in Table 1. In the ANN regression model, 6 hidden layer cells and one output cell have been used.

The block diagram of the proposed ANN regression model given in Fig. 7. The ANN regression model was trained with 1000 epochs and 6 validation checks. Hyperbolic tangent sigmoid transfer function used in training phase. The error function in the neural network used in the regression model is the mean square error (MSE).

The feature vector has been applied as the inputs to ANN to pair off the flame with $\lambda$. Levenberg-Marquardt (LM) and Scaled Conjugate Gradient (SCG) learning approaches have been used to update the weights. The effects of these approaches have examined and evaluated together. It is known that LM approach gives more quick results than SCG approach despite the need for more memory [39]. Within the scope of this study, these approaches have been separately tested for all methods and the results were given in tabular form.

## 5. Experimental results

Different experimental studies have been conducted on the same dataset in order to compare the proposed approaches and similar approaches in the literature. The MSE and R metrics have been used to calculate the regression accuracy of the approaches. R values measure the correlation between outputs and targets. MSE is the average squared difference between outputs and targets. The R and MSE metrics are defined in (11) and (12), respectively.

$$\text{MSE} = \frac{1}{C}\sum_{i=1}^{C}\left(\lambda_i - \lambda_i'\right)^2 \quad (11)$$

$$R = \frac{\sum_{i=1}^{C}(\lambda_i - \bar{\lambda})(\lambda_i' - \bar{\lambda'})}{\sqrt{\sum_{i=1}^{C}\left(\lambda_i - \bar{\lambda}\right)^2 \sum_{i=1}^{C}\left(\lambda_i' - \bar{\lambda'}\right)^2}} \quad (12)$$

Here, $C$ is data count, the value of $\lambda_i$ is the value of the $\lambda$ corresponding to the ith flame (flue gas analyzer measurement). The value $\bar{\lambda}$ is the mean value of all $\lambda_i$ data. The value of $\lambda_i'$ is the $\lambda$ value estimated by the proposed approach for the ith flame image (ANN regression model estimation). The value $\bar{\lambda'}$ is the mean value of all $\lambda_i'$ data.

In the experimental study, firstly, the effect of similarity sum on different channel combinations has been investigated. 16x16 = 256 features have been obtained from each channel and matched to $\lambda$ with the ANN regression model. The obtained results are given in Table 2. All models were executed 10 times and the results given are average values. The highest matching accuracy is achieved by the combination of the R, G and B channels and the selection of SCG ANN method (R = 0.9626). Individual use of color channels reduced matching accuracy. The matching power of channel blue is the best in SCG but poor in LM. It can be generally evaluated the individual usage results that the individual channels have no absolute superiority to each other. In dual channel usage, combination of red and green channels gave the second highest performance. The best result has been obtained from the combination of R, G and B. However, the addition of I to the properties obtained from the combination of R, G and B has been resulted in a decrease in regression performance. Therefore, R, G and B channels have been found to be sufficient for the matching.

In second stage of the experimental study, the effect of Naive Bayes approach was investigated. The experimental procedure tested different channels has been re-applied using the Naive Bayes approach. The results are given in Naive Bayes column of Table 2. It is surprisingly seen that the best matching accuracy is obtained by the combination of R and G channels.

The third experimental study has been carried out to see the effect of the multivariate normal distribution approach on the matching accuracy. Obtained results are given in multivariate normal distribution column of Table 2. The best match accuracy is obtained by using R and G channel together.

The fourth experimental study has been carried out to see the effect of the GMM approach on the matching accuracy. For this process, the accuracy of the channels for different weights was investigated. Obtained results are given in GMM column of Table 2. The best match accuracy is obtained by using R and G channel together. In the incremental weight investigation, the highest performance was found R = 0.9517 accuracy for $w_G = 0.95$ and $w_R = 0.05$.

It is generally seen from Table 2 that the similarity sum method presents the highest correlation in the SCG learning model. As a result of experimental studies, it has been found that the best method for Image-$\lambda$ matching is "Sum of Similarity". To better understand, the results of the proposed camera-based method (blue color) and the conventional gas analyzer (red color) are plotted together (Fig. 8). As shown in the Fig. 8, the values estimated by the proposed system and measured by the flue gas analyzer are shown in the same graph. The obtained results show that system with camera can simulate the gas analysis system result with high accuracy. The reason for this is that the high-resolution color images obtained from the flame image inside the boiler during the combustion process of coal have sufficient information about how the boiler burns.

In the experimental study, finally, the performance of the proposed approach (Sum of Similarity) and the performances of the current approaches (Hue-Hist, Gray-Co-occurrence, Blue-Hist, PCA, RES-Fro-Infinity) in the literature have been examined together. The effect of ANN's SCG and LM learning options on method accuracy was investigated. The matching results of the Image-λ are given in Table 3. Accordingly, the proposed method for both learning methods (SCG or LM) provided the highest accuracy. The results obtained clearly show that the proposed feature methods are insufficient. The main reason for this is that the meaningful features expressing the burning process are not selected and global feature methods are chosen instead of local features approaches.

## 6. Conclusion

The common characteristic of highly efficient and environmentally friend burners is that they have closed-loop control systems that supervise them under ideal combustion conditions. The λ is the most important parameter that characterizes ideal combustion. It also offers a reference value for ideal combustion. The estimation of the λ from the flame images is the basic step for the formation of the advanced combustion control systems. In this study, the existence of a mathematical relationship between λ and flame images has been investigated. The flame images taken with the CCD camera have been vectorized by gridding, modeling and feature extraction. In order to extract the features from the flame images, four different approaches have been studied in addition to the methods available in the literature. The one-dimensional feature vectors obtained by different methods have been matched to the λ values with the help of ANN regression model. The highest matching accuracy of proposed model is achieved with sum of similarity method by the combination of the R, G and B channels and the SCG training method (R = 0.9626). Hue-Histogram method provided the closest accuracy to the proposed method (R = 0.9391). However, it was observed that the feature vector length of the proposed method was much more than the other methods. This means that the processing cost of the proposed method is higher than other methods. However, it is known that the lambda measuring device can provide 1 data per second. This means that the matching time of image-λ must be less than 1 sec. It has been observed that all methods included in the proposed method can perform matching below 1 sec. Thus, the high processing cost of the proposed method did not have any negative impact. Method performances are given comparatively. Experimental results show that proposed prediction schema-based sum of similarity method can provide a quite high accuracy for the estimation of λ. Nut-coal was used in the existing experimental setup. In future studies, a similar experimental setup can be set up for different types of coal and other fuel types under different combustion conditions, and the performance of the proposed method can be tested in different situations.

**Declaration of Competing Interest**

The authors declare that they have no known competing financial interests or personal relationships that could have appeared to influence the work reported in this paper.

**Acknowledgements**

This work was supported by The Scientific and Technological Research Council of Turkey (TUBITAK, Project number: 117M121) and MIMSAN AŞ.

ARTICLE IN PRESS

Estimation of excess air coefficient on coal combustion processes 11[20] J. Chen, L.L.T. Chan, Y.C. Cheng, Gaussian process regression based optimal design of combustion systems using flame images, Appl. Energy 111 (2013) 153–160, https://doi.org/10.1016/j.apenergy.2013.04.036.

tomographic imaging and pyrometric measurement, IST 2012– 2012 IEEE Int. Conf. Imaging Syst. Tech. Proc. (2012) 13–17, https://doi.org/10.1109/IST.2012.6295577.

[21] M.F. Talu, C. Onat, M. Daskin, Prediction of excess air factor in automatic feed coal burners by processing of flame images, Chinese J. Mech. Eng. 30 (3) (May 2017) 722–731, https://doi.org/10.1007/s10033-017-0095-3.

[22] A. González-Cencerrado, B. Peña, A. Gil, Experimental analysis of biomass co-firing flames in a pulverized fuel swirl burner using a CCD based visualization system, Fuel Process. Technol., vol. 130, no. C, pp. 299–310, Feb. 2015, doi: 10.1016/j.fuproc.2014.10.041.

[23] R. Hernández, J. Ballester, Flame imaging as a diagnostic tool for industrial combustion, Combust. Flame 155 (3) (2008) 509–528, https://doi.org/10.1016/j.combustflame.2008.06.010.

[24] Z.-W.W. Jiang, Z.-X.X. Luo, H.-C.C. Zhou, A simple measurement method of temperature and emissivity of coal-fired flames from visible radiation image and its application in a CFB boiler furnace, Fuel 88 (2009) 980–987, https://doi.org/10.1016/j.fuel.2008.12.014.

[25] A. González-Cencerrado, B. Peña, A. Gil, Coal flame characterization by means of digital image processing in a semi-industrial scale PF swirl burner, Appl. Energy 94 (2012) 375–384, https://doi.org/10.1016/j.apenergy.2012.01.059.

[26] S. Golgiyaz, M.F. Talu, C. Onat, Estimation of flue gas temperature by image processing and machine learning methods, Eur. J. Sci. Technol., pp. 283–291, Aug. 2019, doi: 10.31590/ejosat.568348.

[27] H. Zhou, Q. Tang, L. Yang, Y. Yan, G. Lu, K. Cen, Support vector machine based online coal identification through advanced flame monitoring, Fuel vol. 117, no. PARTB (2014) 944–951, https://doi.org/10.1016/j.fuel.2013.10.041.

[28] S. Golgiyaz, M.F. Talu, C. Onat, Artificial neural network regression model to predict flue gas temperature and emissions with the spectral norm of flame image, Fuel 255 (Nov. 2019), https://doi.org/10.1016/j.fuel.2019.115827 115827.

[29] K. Kohse-Hö Inghaus et al, Combustion at the focus: Laser diagnostics and control, Proc. Combust. Inst. 30 (1) (2005) 89–123, https://doi.org/10.1016/j.proci.2004.08.274.

[30] M. Aldén, J. Bood, Z. Li, M. Richter, Visualization and understanding of combustion processes using spatially and temporally resolved laser diagnostic techniques, Proc. Combust. Inst. 33 (1) (2011) 69–97, https://doi.org/10.1016/j.proci.2010.09.004.

[31] C. Zhou, H., Han, An exploratory investigation of the computer-based control of utility coal-fired boiler furnace combustion, J. Eng. Therm. Energy Power, 1994, pp. 111–116.

[32] D. Sun, G. Lu, H. Zhou, Y. Yan, S. Liu, Quantitative assessment of flame stability through image processing and spectral analysis, IEEE Trans. Instrum. Meas. 64 (12) (2015) 3323–3333, https://doi.org/10.1109/TIM.2015.2444262.

[33] Z. Luo, H.C. Zhou, A combustion-monitoring system with 3-D temperature reconstruction based on flame-image processing technique, IEEE Trans. Instrum. Meas. 56 (5) (2007) 1877–1882, https://doi.org/10.1109/TIM.2007.904489.

[34] Z. Luo, F. Wang, H. Zhou, R. Liu, W. Li, G. Chang, Principles of optimization of combustion by radiant energy signal and its application in a 660 MWe down- and coal-fired boiler, Korean J. Chem. Eng. 28 (12) (Dec. 2011) 2336–2343, https://doi.org/10.1007/s11814-011-0098-1.

[35] Z. Guoyi, Q. Jianwei, S. Yipeng, L. Zhonggen, L. Zixue, Z. Huaichun, Experimental detection of radiative energy signal from a supercharged marine boiler and simulation on its application in control of drum water level, Appl. Therm. Eng. 31 (16) (2011) 3168–3175, https://doi.org/10.1016/j.applthermaleng.2011.05.042.

[36] Kenneth C. Weston, Energy Conversion, 1st ed., PWS Publishers, 1992.

[37] W.S. McCulloch, W. Pitts, A logical calculus of the ideas immanent in nervous activity, Bull. Math. Biophys. 5 (4) (Dec. 1943) 115–133, https://doi.org/10.1007/BF02478259.

[38] R.K. Mehra, H. Duan, S. Luo, A. Rao, F. Ma, Experimental and artificial neural network (ANN) study of hydrogen enriched compressed natural gas (HCNG) engine under various ignition timings and excess air ratios, Appl. Energy 228 (Oct. 2018) 736–754, https://doi.org/10.1016/j.apenergy.2018.06.085.

[39] R. Tabbussum, A.Q. Dar, Comparative analysis of neural network training algorithms for the flood forecast modelling of an alluvial Himalayan river, J. Flood Risk Manag. 13 (4) (2020) Dec, https://doi.org/10.1111/jfr3.12656.